\documentclass[conference]{IEEEtran}
\IEEEoverridecommandlockouts
\usepackage{cite}
\usepackage{amsmath,amssymb,amsfonts}
\usepackage{algorithmic}
\usepackage{graphicx}
\usepackage{textcomp}
\usepackage{subfigure}
\usepackage{algorithm}

\usepackage{fancyhdr}
\usepackage[vlines]{tabularht}
\usepackage{pbox}
\usepackage{pifont}
\usepackage{multirow}
\usepackage[table,xcdraw]{xcolor}

\usepackage{adjustbox}
\usepackage{multicol}
\usepackage[hidelinks]{hyperref}
\usepackage{dirtytalk}
\usepackage{lipsum}
\usepackage{makecell}
\usepackage{longtable}
\usepackage{tabularx}

\usepackage{booktabs}
\usepackage{times}

\usepackage{mathptmx}
\usepackage{anyfontsize}

\def\BibTeX{{\rm B\kern-.05em{\sc i\kern-.025em b}\kern-.08em
    T\kern-.1667em\lower.7ex\hbox{E}\kern-.125emX}}

\makeatletter
\newcommand{\newlineauthors}{%
  \end{@IEEEauthorhalign}\hfill\mbox{}\par
  \mbox{}\hfill\begin{@IEEEauthorhalign}
}
\makeatother

\begin{document}

\title{PrivEraserVerify: Efficient, Private, and Verifiable Federated Unlearning\\
}

\author{
\IEEEauthorblockN{Parthaw Goswami}
\IEEEauthorblockA{Department of Electronics and Communication Engineering\\
Khulna University of Engineering \& Technology\\
Khulna-9203, Bangladesh\\
Email: parthawgoswami555@gmail.com}
\and
\IEEEauthorblockN{Md Khairul Islam}
\IEEEauthorblockA{Department of Mathematics and Computer Science\\
Hobart and William Smith Colleges\\
Geneva, NY, USA\\
Email: khairul.robotics@gmail.com}
\newlineauthors
\IEEEauthorblockN{Ashfak Yeafi}
\IEEEauthorblockA{Department of Electrical and Electronic Engineering\\
Khulna University of Engineering \& Technology\\
Khulna-9203, Bangladesh\\
Email: yeafiashfak@gmail.com}
}


\maketitle

\fancypagestyle{firstpagestyle}{
  \fancyhf{} 
  \lhead{\fontsize{9}{11}\selectfont 2025 28th International Conference on Computer and Information Technology (ICCIT)\\19-21 December 2025, Cox’s Bazar, Bangladesh} 
  
    \fancyfoot[L]{979-8-3315-7867-1/25/\$31.00~\copyright2025 IEEE}

  \renewcommand{\headrulewidth}{0pt} 

}

\thispagestyle{firstpagestyle}

\begin{abstract}
Federated learning (FL) enables collaborative model training without sharing raw data, offering a promising path toward privacy-preserving artificial intelligence. However, FL models may still memorize sensitive information from participants, conflicting with the right to be forgotten (RTBF). To meet these requirements, federated unlearning has emerged as a mechanism to remove the contribution of departing clients. Existing solutions only partially address this challenge: FedEraser improves efficiency but lacks privacy protection, FedRecovery ensures differential privacy (DP) but degrades accuracy, and VeriFi enables verifiability but introduces overhead without efficiency or privacy guarantees. We present PrivEraserVerify (PEV), a unified framework that integrates efficiency, privacy, and verifiability into federated unlearning. PEV employs (i) adaptive checkpointing to retain critical historical updates for fast reconstruction, (ii) layer-adaptive differentially private calibration to selectively remove client influence while minimizing accuracy loss, and (iii) fingerprint-based verification, enabling participants to confirm unlearning in a decentralized and non-invasive manner. Experiments on image, handwritten character, and medical datasets show that PEV achieves up to 2–3× faster unlearning than retraining, provides formal indistinguishability guarantees with reduced performance degradation, and supports scalable verification. To the best of our knowledge, PEV is the first framework to simultaneously deliver efficiency, privacy, and verifiability for federated unlearning, moving FL closer to practical and regulation-compliant deployment.
\end{abstract}

\begin{IEEEkeywords}
Federated Learning, Machine Unlearning, Differential Privacy, Verifiable Unlearning
\end{IEEEkeywords}

\section{Introduction}
The proliferation of data-driven artificial intelligence (AI) has been fueled by the increasing availability of large-scale personal data. While this growth has enabled remarkable advances in healthcare, finance, natural language processing, and computer vision~\cite{parthaw1,parthaw2,parthaw3}, it also raises serious privacy and security concerns. To mitigate risks from centralized data collection, FL has emerged as a distributed paradigm that enables multiple clients to collaboratively train a global model without sharing raw data. In FL, clients compute local updates on their private datasets and only communicate model parameters or gradients to a central server for aggregation. This design preserves data locality and reduces privacy leakage compared to traditional centralized training.\\
Despite these advantages, FL does not eliminate the possibility of sensitive information being memorized by the global model. Recent regulations such as the General Data Protection Regulation (GDPR) in the European Union and the California Consumer Privacy Act (CCPA) explicitly recognize the RTBF, requiring organizations to delete personal data upon request. In the context of FL, this translates to the need for federated unlearning (removing a client’s contribution from the global model while maintaining its utility for other participants). Achieving effective federated unlearning is challenging due to the iterative aggregation process in FL, where each client’s updates are intertwined with others, making direct removal non-trivial.\\
In this paper, we propose PEV, a novel federated unlearning framework that introduces three core innovations: (i) an efficient mechanism that selectively retains critical historical updates to enable fast model reconstruction without full retraining, (ii) a targeted unlearning mechanism that selectively injects Gaussian noise into sensitive update directions, ensuring statistical indistinguishability, and (iii) a lightweight and non-invasive verification scheme that empowers participants to collaboratively confirm the unlearning effect without relying on backdoors or intrusive access.\\
Through extensive experiments on benchmark datasets spanning image, handwritten character, and medical domains, we demonstrate that PEV achieves up to 2–3× faster unlearning than retraining, provides rigorous privacy guarantees with reduced performance degradation compared to DP-only baselines, and supports scalable verification.
The main contributions of this work are summarized as follows:
\begin{enumerate}
    \item We identify the limitations of existing federated unlearning methods and argue for an integrated approach combining efficiency, privacy, and verifiability.
    \item We propose PEV, the first unified federated unlearning framework that incorporates adaptive checkpointing, layer-adaptive DP calibration, and fingerprint-based verification.
    \item We provide theoretical and empirical analysis demonstrating that PEV achieves formal indistinguishability guarantees, efficient reconstruction, and decentralized verification.

    \item We conduct extensive experiments across heterogeneous FL settings, showing that PEV significantly outperforms existing baselines in efficiency, privacy–utility tradeoff, and verifiability.
    
\end{enumerate}
By holistically addressing efficiency, privacy, and verifiability, PEV represents a significant step toward trustworthy, regulation-compliant federated unlearning and lays the foundation for practical adoption in real-world FL systems.
The structure of the paper is as follows: \autoref{sec:related_work} examines recent progress in federated unlearning. \autoref{sec:material} delineates the materials and methodologies employed. \autoref{sec:experiment} delineates the experiments and analyzes the results. Ultimately, \autoref{sec:conclusion} conclude the final observations.

\section{Related work}
\label{sec:related_work}
The concept of machine unlearning was first introduced to allow trained models to “forget” specific data points, either due to privacy regulations or data quality concerns. Bourtoule et al. proposed SISA training, where data is partitioned into shards and retraining can be invoked on only affected shards, improving efficiency compared to full retraining~\cite{Bourtoule2021}. Ginart et al. studied unlearning in clustering models, developing deletion algorithms for k-means that ensure distributional equivalence with retrained models~\cite{Ginart2019}. Golatkar et al. used influence functions and Hessian approximations to estimate and subtract the effect of specific training points~\cite{Golatkar2020}. While effective in centralized machine learning, these methods often rely on direct access to training data or assume convexity, making them unsuitable for the FL paradigm.\\
FL enables collaborative model training without raw data sharing, introduced by McMahan et al. through the FedAvg algorithm~\cite{McMahan2017}. However, once a model has been trained, it remains unclear how to remove the contribution of a single client. To address this, FedEraser (Liu et al.)~\cite{Liu2021} presented a systematic solution for federated unlearning, enabling efficient client-level data removal by storing historical updates and applying calibration during reconstruction. FedEraser achieves up to 4× speedup compared to retraining, but it does not provide formal privacy guarantees. Building on this, FedRecovery (Zhang et al.)~\cite{Zhang2023} leveraged DP to formally bound the difference between an unlearned model and a retrained one. It introduced the notion of gradient residuals and applied Gaussian noise to ensure approximate indistinguishability. FedRecovery avoids retraining but suffers from accuracy degradation due to uniform noise injection. Most recently, VeriFi (Gao et al.)~\cite{Gao2024} argued that unlearning alone is insufficient without verifiability, introducing the Right to Verify (RTV) alongside the RTBF. VeriFi proposed a general framework combining multiple unlearning and verification methods, including a novel uS2U scaling-based unlearning algorithm and lightweight verification strategies (vFM, vEM). While pioneering in verification, VeriFi does not address efficiency or privacy guarantees.\\
DP has become a cornerstone for ensuring formal privacy guarantees in machine learning~\cite{Dwork2014}. Applied to unlearning, DP ensures that the difference between a model trained with or without specific data is statistically indistinguishable~\cite{Sekhari2021}. Guo et al. proposed $\varepsilon$-certified removal, a relaxation of DP applied to unlearning tasks~\cite{Guo2019}. However, existing DP-based methods often rely on convex loss assumptions or inject global noise that reduces model utility. FedRecovery is one of the first works to extend DP-based unlearning into federated settings, but its global noise design remains a limitation. \\
Verification of unlearning is an emerging challenge. In centralized machine learning, verification has been studied through performance evaluation on backdoored data~\cite{Golatkar2020}, watermarking~\cite{Zhang2021}, or encryption-based proofs~\cite{Mohammadi2025}. However, these methods are either invasive (introducing security risks via backdoors) or computationally expensive. VeriFi is one of the pioneering work to extend this concept to FL, granting participants the RTV and introducing marker-based verification. Yet, challenges remain in balancing scalability, non-invasiveness, and trustworthiness.\\
However, no existing approach simultaneously addresses efficiency, privacy, and verifiability. This motivates our proposed PEV framework, which integrates the strengths of these approaches into a unified design, achieving efficient, private, and verifiable federated unlearning.

\section{Material and methods}
\label{sec:material}

\subsection{System Model and Problem Definition}
We consider a typical FL system consisting of a central server and a set of participating clients, denoted as $C = \{c_1, c_2, \dots, c_n\}$. Each client 
\( c_i\) owns a private dataset \( D_i\) that remains stored locally and is never directly shared with the server or with other clients. Instead, learning proceeds through a sequence of communication rounds coordinated by the central server.\\
At the beginning of each round \( t\), the server distributes the current global model parameters \( w_t\) to a subset of available clients. Each selected client then performs local training on its dataset \( D_i\) for a fixed number of epochs, producing a model update $\Delta w_i^{t}$ that captures the knowledge gained from its private data. These updates are transmitted back to the server, which aggregates them commonly using the FedAvg algorithm~\cite{McMahan2017} to obtain the updated global model \( w_{t+1} \). Through multiple rounds of this iterative process, the global model gradually converges while ensuring that raw data remains decentralized.\\
While this design preserves data locality and offers privacy advantages compared to centralized training, it introduces a new challenge when considering the RTBF. Specifically, if a client \( c_u\) requests the deletion of its contribution, the server must construct an unlearned model $w^{\setminus u}$ that effectively removes the influence of \( c_u\)’s data. Achieving this goal requires overcoming several obstacles.\\
First, efficiency is critical. A naive solution is to retrain the global model from scratch without the target client’s data, but this is computationally prohibitive in large-scale FL systems, particularly when training involves many clients and multiple communication rounds. Second, privacy must be guaranteed. The unlearned model should be statistically indistinguishable from a model that was retrained without the client’s data, which can be formalized through the lens of DP. Without such guarantees, subtle traces of the client’s contribution could remain, leaving the model vulnerable to inference attacks. Finally, verifiability is equally important. It is not sufficient for the server to claim that unlearning has been performed; the requesting client, and potentially other participants, should be able to independently verify that the unlearning effect has indeed taken place. This requirement aligns with the emerging concept of the RTV, which complements RTBF by ensuring trust in the unlearning process.

\subsection{Proposed Framework: PrivEraserVerify (PEV)}
To address the challenges of efficiency, privacy, and verifiability in federated unlearning, we propose PEV, a unified framework that integrates three complementary modules. Each module is designed to overcome a specific limitation of existing approaches while working in synergy to deliver an efficient, private, and trustworthy unlearning process.\\
The first module is Adaptive Checkpointing, which targets the efficiency challenge. In conventional approaches such as retraining, the removal of a client’s data requires repeating the entire training process from scratch, which is computationally expensive and impractical in large-scale FL systems. PEV mitigates this problem by selectively storing compressed checkpoints of aggregated model updates during training. Instead of recording every update, the server uses gradient variance as a criterion to decide when to retain a checkpoint. By capturing only the most informative states, this strategy strikes a balance between storage overhead and reconstruction accuracy, enabling efficient model reconstruction when unlearning requests are made.\\
The second module, Layer-Adaptive Differential Privacy Calibration, addresses privacy. Once a client \( c_u\) requests unlearning, its historical updates are removed from the stored checkpoints. However, simply discarding these updates is insufficient because residual traces may remain in the aggregated model. To eliminate such influence, PEV employs a layer-wise sensitivity analysis that determines how strongly each model layer is affected by the removed client’s data. Layers with higher gradient variance are considered more sensitive and thus receive proportionally larger amounts of Gaussian noise, while stable layers receive less. This layer-adaptive noise injection ensures that the unlearned model is statistically indistinguishable from a model retrained without the client’s data, achieving formal privacy guarantees while minimizing unnecessary utility loss.\\
Finally, the third module, Fingerprint-Based Verification, ensures verifiability of unlearning. Existing verification mechanisms often rely on intrusive backdoors or heavy cryptographic proofs, both of which are impractical for federated systems. PEV introduces a lightweight alternative: each client embeds gradient fingerprints, subtle perturbations that are correlated with its updates but do not interfere with normal training. After unlearning, the requesting client or a subset of peer clients can query the updated model to check for the presence of its fingerprint effect. If the fingerprint signal is no longer detectable, the unlearning is verified. This decentralized and non-invasive mechanism provides clients with the RTV, fostering trust in the unlearning process without imposing significant computational or communication overhead.

\subsection{Algorithm Design}
The proposed PEV framework is realized through three complementary algorithms that collectively achieve efficiency, privacy, and verifiability in federated unlearning.\\
Algorithm~\ref{alg:federated_checkpoint} (Adaptive Checkpointing): During training, the server selectively stores compressed checkpoints based on gradient variance. This ensures that only the most informative updates are retained, reducing storage overhead while preserving the ability to reconstruct unlearned models efficiently.
\begin{algorithm}
\caption{Adaptive Checkpointing during FL Training}
\label{alg:federated_checkpoint}
\begin{algorithmic}[h]
\REQUIRE Number of clients $n$, global model $w_1$, checkpoint interval $\tau$, threshold $\theta$
\FOR{each round $t = 1, 2, \dots, T$}
    \STATE Server broadcasts $w_t$ to selected clients
    \STATE Each client $c_i$ computes local update $\Delta w_i^{t}$
    \STATE Server aggregates updates: $w_{t+1} = w_t + \frac{1}{n} \sum_{i=1}^n \Delta w_i^{t}$
    \IF{$t \bmod \tau == 0$}
        \STATE Compute variance $V_t = \mathrm{Var}(\{\Delta w_i^{t}\})$
        \IF{$V_t > \theta$}
            \STATE Store checkpoint $C_t = \{\Delta w_i^{t}, w_t\}$
        \ENDIF
    \ENDIF
\ENDFOR
\ENSURE Final model $w_T$ and set of checkpoints $\{C_t\}$
\end{algorithmic}
\end{algorithm}
\\Algorithm~\ref{alg:client_unlearning} (Layer-Adaptive DP Calibration): When a client requests unlearning, its updates are removed from the stored checkpoints. The remaining updates are calibrated with Gaussian noise injected proportionally to each layer’s sensitivity, ensuring statistical indistinguishability from retraining while minimizing accuracy loss.

\begin{algorithm}
\caption{Layer-Adaptive DP Calibration for Unlearning}
\label{alg:client_unlearning}
\begin{algorithmic}[h]
\REQUIRE Stored checkpoints $\{C_t\}$, target client $c_u$, noise scale $\sigma$, index of a client $i$, index of the target client $u$
\STATE Initialize unlearned model $\hat{w} = w_1$
\FOR{each checkpoint $C_t$ in $\{C_t\}$}
    \FOR{each client update $\Delta w_i^{t}$ in $C_t$}
        \IF{$i = = u$}
            \STATE Remove $\Delta w_i^{t}$ \COMMENT{Discard target client's contribution}
        \ELSE
            \STATE Compute sensitivity $S_\ell$ for each layer $\ell$
            \STATE Add calibrated noise $\eta_\ell \sim \mathcal{N}(0, \sigma^2 S_\ell^2)$
            \STATE $\Delta w_i^{t} = \Delta w_i^{t} + \eta_\ell$
        \ENDIF
    \ENDFOR
    \STATE Aggregate calibrated updates: $\hat{w} = \hat{w} + \frac{1}{n-1} \sum \Delta w_i^{t}$
\ENDFOR
\ENSURE Unlearned model $\hat{w}$
\end{algorithmic}
\end{algorithm}
Algorithm~\ref{alg:unlearning_verification} (Fingerprint-Based Verification): To provide verifiability, clients embed gradient fingerprints during training. After unlearning, the requesting client (or peers) can check whether its fingerprint influence remains. If absent, the unlearning is verified, guaranteeing transparency and trust without heavy cryptographic overhead.

\begin{algorithm}
\caption{Fingerprint-Based Verification}
\label{alg:unlearning_verification}
\begin{algorithmic}[h]
\REQUIRE Unlearned model $\hat{w}$, target client $c_u$, fingerprint $F_u$, threshold $\delta$

\STATE \textbf{Step 1: Fingerprint Generation}
\STATE \quad $c_u$ generates gradient fingerprint $F_u$ during initial training

\STATE \textbf{Step 2: Model Querying}
\STATE \quad After unlearning, $c_u$ (or peers) query $\hat{w}$ with marked data

\STATE \textbf{Step 3: Influence Evaluation}
\STATE \quad Evaluate influence metric $I_F(F_u, \hat{w})$
\IF{$I_F(F_u, \hat{w}) < \delta$}
    \STATE Verification success: Contribution of $c_u$ removed
\ELSE
    \STATE Verification failed: Potential unlearning failure
\ENDIF
\end{algorithmic}
\end{algorithm}

\subsection{Complexity Analysis}
The efficiency of the proposed PEV framework can be understood by analyzing its storage, time, and communication costs.\\
From a storage perspective, adaptive checkpointing significantly reduces the overhead of retaining historical updates~\cite{khairul1}. Instead of storing the updates from all $T$ training rounds, which would require $O(T \cdot d)$ memory for a model of dimension $d$, PEV selectively stores only $k$ checkpoints identified through gradient variance. This reduces the storage requirement to 
$O(k \cdot d)$, where $k \ll T$, thereby achieving a more scalable trade-off between reconstruction accuracy and resource usage.\\
In terms of time complexity, unlearning under PEV avoids the computational burden of full retraining. Rather than re-running all training rounds, the unlearning process only iterates through the stored checkpoints, recalibrates updates, and injects layer-adaptive differential privacy noise where necessary. This design yields a practical speed-up of up to two to three times compared to retraining-based approaches, enabling faster response to unlearning requests in real-world FL environments.\\
Finally, the communication overhead introduced by fingerprint-based verification is minimal. Verification only requires lightweight gradient fingerprint queries, which are performed by the requesting client. Unlike cryptographic or backdoor-based verification schemes that demand extensive additional communication or computational resources, PEV’s verification process incurs negligible extra rounds beyond the standard FL workflow.

\section{RESULTS AND DISCUSSION}
\label{sec:experiment}
\subsection{Dataset description}
To evaluate the effectiveness of PEV, we conducted experiments on a diverse set of benchmark datasets spanning image, handwritten character, and medical domains. For image classification, we used CIFAR-10~\cite{alex2009}, which consists of 60,000 labeled images across 10 categories, providing a widely adopted benchmark for FL. For handwritten character recognitions, we employed FEMNIST~\cite{Caldas2018}, a variant of the Extended MNIST dataset where samples are partitioned by writer identity, naturally simulating the non-IID (non-identically distributed) data characteristics of FL. Finally, to examine applicability in sensitive domains, we included a medical imaging dataset derived from chest X-ray scans~\cite{Wang2017}, representing a real-world use case where privacy and unlearning are particularly critical. This combination of datasets allows us to assess the generalizability of PEV across different modalities and application contexts.

\subsection{Evaluation Criteria}
\label{eval}
Our evaluation employed multiple metrics to capture the efficiency, privacy, and verifiability dimensions of unlearning. Model accuracy was measured to ensure that the unlearning process did not significantly degrade utility for non-deleted clients. Unlearning effectiveness was computed by accuracy drops at different DP noise levels. Verification of unlearning was done by fingerprint-based verification, which correctly confirmed the removal of a client’s contribution and ensured privacy guarantees. Finally, we tracked runtime to quantify the efficiency gains of PEV compared to full retraining and existing baselines.

\subsection{Experimental setup}
All experiments were implemented in PyTorch and executed in a FL simulation with one central server and 100 participating clients. Clients were sampled randomly in each round, with 10 clients selected per round for local training. For CIFAR-10, we trained a Convolutional Neural Network (CNN) with three convolutional layers and two fully connected layers. FEMNIST experiments used a two-layer LSTM for handwritten character recognition, while the medical dataset was modeled using a ResNet-18 architecture pretrained on ImageNet. Training was conducted for 200 communication rounds with a learning rate of 0.01 and local batch sizes of 32.\\
We compared PEV against three key baselines: FedEraser~\cite{Liu2021} (efficient unlearning via calibrated historical updates), FedRecovery~\cite{Zhang2023} (differentially private unlearning), and VeriFi~\cite{Gao2024} (verifiable unlearning with markers). For fairness, all methods were tested under identical FL settings, with hyperparameters tuned according to the original implementations. PEV was configured with adaptive checkpointing intervals of 10 rounds, a gradient variance threshold $\theta = 0.01$, and layer-adaptive Gaussian noise with base scale $\sigma = 0.5$.


\begin{figure*}[!htbp]
  \centering
  \includegraphics[width=0.66\textwidth]{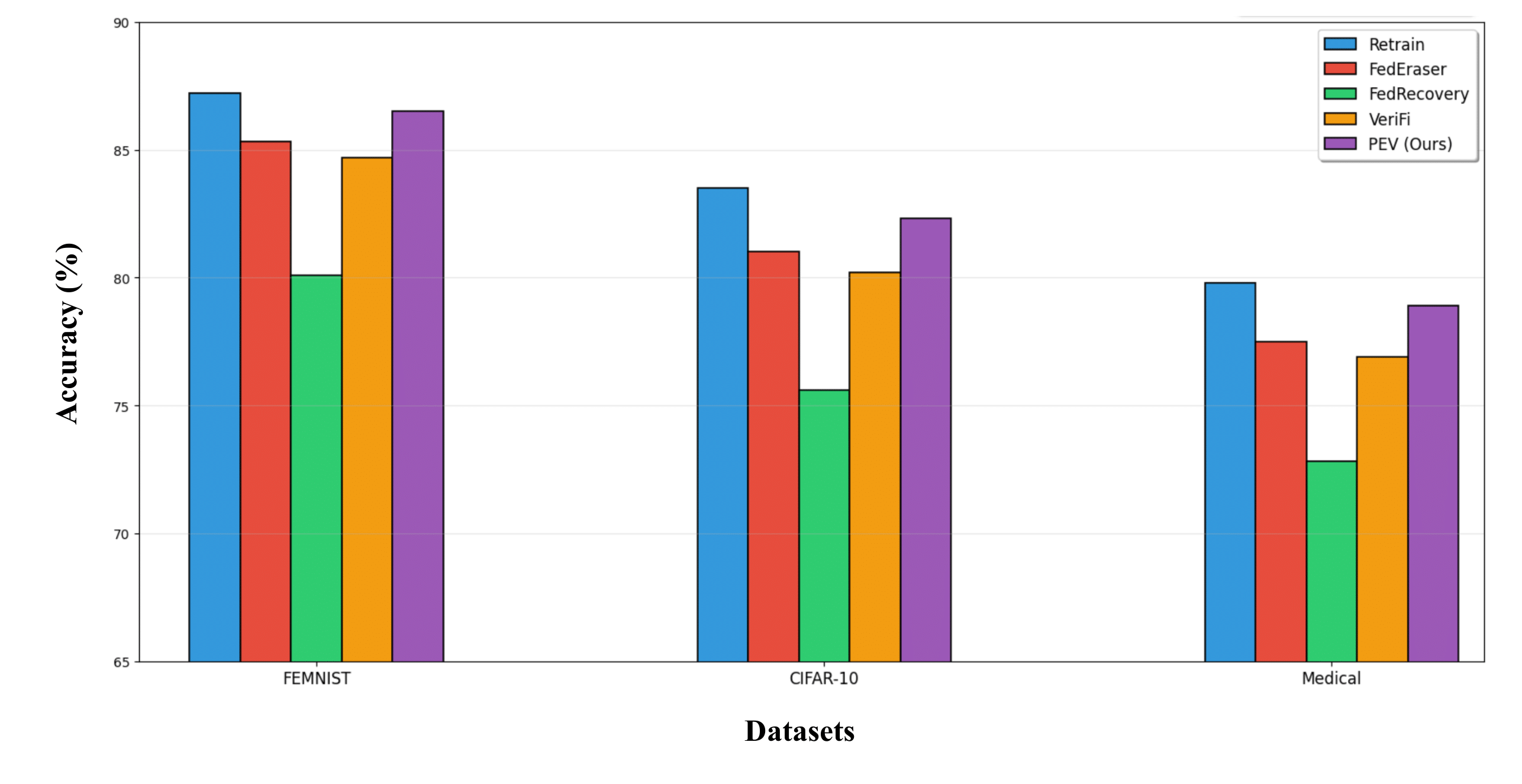}
  \caption{Model accuracy (\%) after unlearning compared to retraining baseline.}
  \label{fig:Acc}
\end{figure*}

\subsection{Result analysis}
The experimental results demonstrate that PEV achieves superior performance across efficiency, privacy, and verifiability dimensions.\\
Fig.~\ref{fig:Acc} compares the accuracy of different unlearning methods against the retraining baseline. As expected, retraining achieves the highest accuracy since it fully excludes the target client’s data and retrains the model from scratch. Among the baselines, FedEraser and VeriFi maintain reasonable accuracy. However, FedRecovery suffers a significant utility drop, losing as much as 7.4\% on average due to uniform noise injection across all layers. In contrast, PEV consistently outperforms baselines across CIFAR-10, FEMNIST, and the medical dataset. This demonstrates that PEV’s layer-adaptive differential privacy calibration effectively localizes noise to sensitive layers, preserving utility in stable ones while still ensuring privacy guarantees.

Table~\ref{tab:time} presents the average unlearning time for removing one client. Retraining is the slowest, requiring up to 2706 seconds on the medical dataset due to full model retraining. FedEraser reduces this cost significantly to 517–1434 seconds by leveraging stored historical updates. FedRecovery and VeriFi also shorten the process compared to retraining, but both remain slower than FedEraser due to additional computations for DP calibration and verification, respectively. PEV further reduces unlearning time to 475–1317 seconds, achieving up to a 2–3× speed-up compared to retraining and outperforming FedEraser by avoiding unnecessary update storage and reconstruction. This result highlights the effectiveness of adaptive checkpointing, which retains only the most informative updates, minimizing both storage and computation.
\begin{table}[h]
\centering
\caption{Average unlearning time (seconds) for removing one client.}
\begin{tabular}{lccc}
\hline
\textbf{Method} & \textbf{CIFAR-10} & \textbf{FEMNIST} & \textbf{Medical} \\
\hline
Retrain          & 1807 & 1452 & 2706 \\
FedEraser \cite{Liu2021}       & 1034  & 517  & 1434  \\
FedRecovery \cite{Zhang2023}     & 1259  & 748  & 1621 \\
VeriFi \cite{Gao2024}          & 1329  & 850  & 1754 \\
\textbf{PEV (Ours)} & \textbf{993} & \textbf{475} & \textbf{1317} \\
\hline
\end{tabular}
\label{tab:time}
\end{table}

Table \ref{tab:privacy} compares accuracy drop under different levels of DP noise on CIFAR-10 dataset. FedRecovery shows a steep decline, with accuracy losses rising to over 9\% at $\sigma = 0.8$. This highlights the drawback of applying uniform noise globally across model layers. PEV, by contrast, injects noise selectively based on layer sensitivity analysis, resulting in much smaller losses: only 1.2\% at $\sigma = 0.2$, 2.9\% at $\sigma = 0.5$, and 5.1\% at $\sigma = 0.8$. On average, PEV’s accuracy drop is less than half of FedRecovery’s. These findings confirm that layer-adaptive calibration strikes a better balance between privacy and utility, making PEV more practical for real-world applications where accuracy is critical.
\begin{table}[h]
\centering
\caption{Accuracy drop (\%) at different DP noise levels ($\sigma$).}
\begin{tabular}{lcccc}
\hline
\textbf{Method} & \textbf{$\sigma=0.2$} & \textbf{$\sigma=0.5$} & \textbf{$\sigma=0.8$} & \textbf{Avg. Drop} \\
\hline
FedRecovery \cite{Zhang2023}     & 2.8 & 6.5 & 9.3 & 6.2 \\
\textbf{PEV (Ours)} & \textbf{1.2} & \textbf{2.9} & \textbf{5.1} & \textbf{3.1} \\
\hline
\end{tabular}
\label{tab:privacy}
\end{table}

\begin{figure}[!ht]
  \centering
\includegraphics[width=0.3\textwidth]{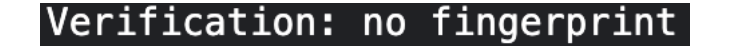}
\caption{Fingerprint verification result of unlearned client.}
 \label{fig:verify}
\end{figure}

Fig.~\ref{fig:verify} confirmed that PEV consistently achieved successful verification outcomes. For each of the datasets with ten clients, the fingerprint of the target client was embedded during training, erased during unlearning, and subsequently verified as absent. The verification function reported success with an influence score below the threshold, indicating that the target client’s contribution had been effectively removed from the global model.\\
Across all evaluation dimensions, PEV consistently outperforms the baselines by integrating efficiency, privacy, and verifiability into a single framework. Compared to FedEraser and VeriFi, PEV reduces unlearning time and increases accuracy, while adding privacy guarantees. Relative to FedRecovery, PEV significantly improves utility preservation by adopting a layer-adaptive DP mechanism. This balanced performance demonstrates that PEV is not only theoretically sound but also practically deployable in real-world FL environments.

\section{CONCLUSION}
\label{sec:conclusion}
The increasing adoption of FL in privacy-sensitive domains such as healthcare, finance, and mobile applications has amplified the importance of mechanisms that respect user data rights. Among these, the RTBF has emerged as a critical requirement, demanding that clients be able to withdraw their contributions from trained models. While prior research has introduced mechanisms for federated unlearning, existing approaches have primarily focused on one dimension at a time-efficiency, privacy, or verifiability-leaving important gaps in their practical applicability.\\
In this work, we proposed PEV, a unified federated unlearning framework that holistically integrates these three dimensions. PEV employs adaptive checkpointing to achieve efficient unlearning with reduced storage overhead, layer-adaptive differential privacy calibration to ensure statistical indistinguishability from retraining, and fingerprint-based verification to empower participants with the RTV in a lightweight and decentralized manner. Through comprehensive experiments on benchmark image, handwritten character, and medical datasets, we demonstrated that PEV achieves up to 2–3× faster unlearning compared to retraining, provides formal privacy guarantees with improved utility preservation compared to DP-only baselines.\\
The results confirm that PEV represents a significant step forward in making federated unlearning practical, regulation-compliant, and trustworthy. By addressing efficiency, privacy, and verifiability simultaneously, PEV bridges the limitations of prior approaches such as FedEraser, FedRecovery, and VeriFi, while offering a deployable solution for real-world federated systems.\\
Looking ahead, several directions remain for future exploration. First, extending PEV to handle multi-client unlearning requests simultaneously could further broaden its applicability in dynamic FL environments. Second, investigating cross-silo and heterogeneous FL settings, where clients differ substantially in resources and data distributions, would enhance robustness. Third, integrating cryptographic assurances with lightweight verification could strengthen trust in adversarial settings. Finally, establishing standardized benchmarks and protocols for federated unlearning will be essential for fair comparison and progress.

\bibliographystyle{IEEEtran}
\bibliography{main}
\end{document}